\def\year{2018}\relax
\documentclass[letterpaper]{article} 
\usepackage{aaai18}  
\usepackage{times}  
\usepackage{helvet}  
\usepackage{courier}  
\usepackage{url}  
\usepackage{graphicx}  
\usepackage{color}
\usepackage{amsfonts}
\usepackage{amsmath}
\usepackage{booktabs}
\usepackage[table]{xcolor}
\definecolor{mygray2}{gray}{.75}
\def\ttt{\multicolumn{1}{>{\columncolor{mygray2}[1.0\tabcolsep]}c}}
\def\tt{\multicolumn{1}{>{\columncolor{mygray2}[1.0\tabcolsep]}c|}}

\begin{document}
%
\title{Transferable Semi-supervised Semantic Segmentation}
\author{Huaxin Xiao\textsuperscript{1,2}, Yunchao Wei\textsuperscript{3}, Yu Liu\textsuperscript{1}, Maojun Zhang\textsuperscript{1}, Jiashi Feng\textsuperscript{2}\\
\textsuperscript{1}Department of System Engineering, National University of Defense Technology\\
\textsuperscript{2}Department of ECE, National University of Singapore\\
\textsuperscript{3}Beckman Institute, University of Illinois at Urbana-Champaign\\
\{xiaohuaxin, jasonyuliu, mjzhang\}@nudt.edu.cn,
wychao1987@gmail.com, elefjia@nus.edu.sg
}
\maketitle
\begin{abstract}
The performance of deep learning based semantic segmentation models heavily depends on sufficient data with careful annotations. However, even the largest public datasets only provide samples with pixel-level annotations for rather limited semantic categories. Such data scarcity critically limits scalability and applicability of semantic segmentation models in real applications. In this paper, we propose a novel transferable semi-supervised semantic segmentation model that can transfer the learned segmentation knowledge from a few strong categories with pixel-level annotations to unseen weak categories with only image-level annotations, significantly broadening the applicable territory of deep segmentation models. In particular, the proposed model consists of two complementary and learnable components: a Label transfer Network (L-Net) and a Prediction transfer Network (P-Net). The L-Net learns to transfer the segmentation knowledge from strong categories to the images in the weak categories and produces coarse pixel-level semantic maps, by effectively exploiting the similar appearance shared across categories. Meanwhile, the P-Net tailors the transferred knowledge through a carefully designed adversarial learning strategy and produces refined segmentation results with better details. Integrating the L-Net and P-Net achieves $96.5\%$ and $89.4\%$ performance of the fully-supervised baseline using $50\%$ and $0\%$ categories with pixel-level annotations respectively on PASCAL VOC 2012. With such a novel transfer mechanism, our proposed model is easily generalizable to a variety of new categories, only requiring image-level annotations, and offers appealing scalability in real applications.
\end{abstract}

\section{Introduction}
Fully-supervised deep learning algorithms for semantic segmentation ~\cite{long2015fully,chen2014semantic,pan2017fully} generally demand a large amount of high-quality pixel-level annotation. However, such annotation is only available for a small number of categories to date, e.g., 20 categories in PASCAL VOC 2012~\cite{2010-pascal} and 80 categories in MS-COCO~\cite{lin2014microsoft}. Scarcity of annotated data severely limits the deployment of advanced segmentation models in real applications. The semi-supervised learning based semantic segmentation models are developed to provide an alternative for comparable segmentation quality with less annotation cost.

\begin{figure}[t]
	\centering
	\includegraphics[width=0.99\linewidth]{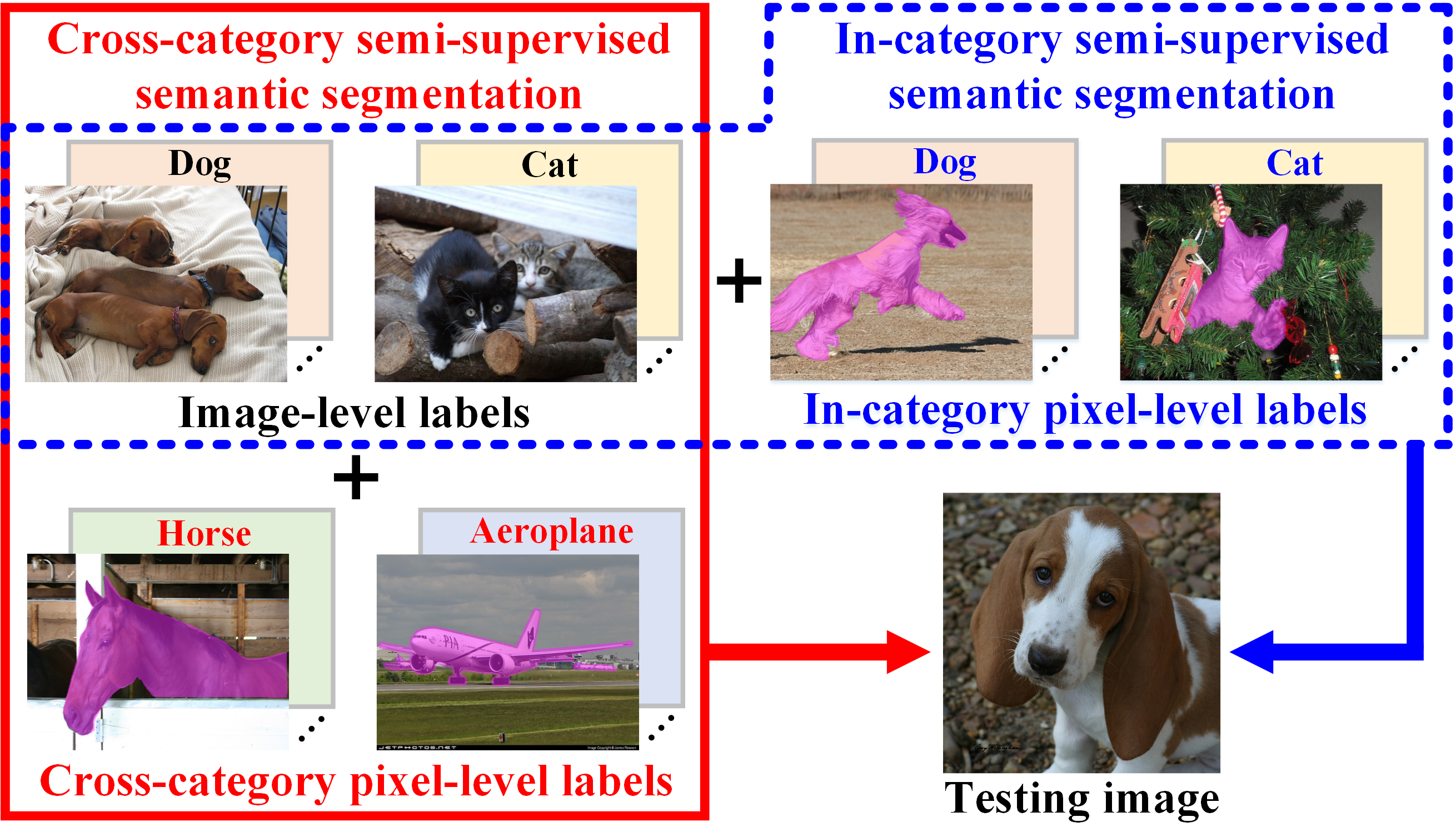}
	\caption{Illustration on two different settings for semi-supervised image semantic segmentation: conventional In-category Semi-supervised Semantic Segmentation (I3S) and the novel Cross-category Semi-supervised Semantic Segmentation (C3S) (which we consider in this work). Different from the I3S problem where each category (e.g., {\ttfamily{dog}}) has a few in-category pixel-level annotations as well as considerable image-level labels, we introduce a more general and realistic C3S problem where some categories (e.g., {\ttfamily{horse}} and {\ttfamily{aeroplane}}) have pixel-level annotations and some other categories to segment (e.g., {\ttfamily{dog}} and {\ttfamily{cat}}) only have image-level labels. The C3S problem is more challenging and requires the segmentation model to have a strong transferable learning ability. Best viewed in color.}
	\label{fig1}
\end{figure}

In the setting of conventional semi-supervised semantic segmentation~\cite{papandreou2015weakly,hong2015decoupled}, namely the In-category Semi-supervised Semantic Segmentation (I3S) as illustrated in the top panel of Figure~\ref{fig1}, each category in the training set must be provided with a few pixel-level annotations as well as considerable image-level annotations. 
This setting however deviates from real applications because a new introduced category still requires extra efforts on re-labeling the category samples. This scheme thus becomes impractical for dealing with hundreds of thousands of categories. For instance, over 20,000 categories are included in ImageNet~\cite{russakovsky2015imagenet} and people can recognize much more categories. 

To mitigate such a gap and essentially enhance scalability and applicability of segmentation models, in this work we introduce a more general learning scheme of semi-supervised semantic segmentation, i.e., the Cross-category Semi-supervised Semantic Segmentation (C3S), as illustrated in the left panel of Figure~\ref{fig1}. Within C3S scheme, different categories have supervision at different levels, or more concretely some categories have pixel-level annotations (called ``strong'' categories) and some only have class labels (called ``weak'' categories). More importantly, there is no overlap between the strong and weak categories.

To solve C3S induced problems, the key point lies in how to effectively learn and transfer re-usable knowledge from strong categories to the segmentation of weak categories. To this end, we develop a novel transferable semi-supervised semantic segmentation model. It contains two complementary components, i.e., a Label transfer Network (L-Net) and a Prediction transfer Network (P-Net), to transfer and adapt the learned segmentation knowledge from strong categories to the weak ones. More concretely, the L-Net learns the segmentation knowledge from strong categories explicitly at first and then transfers the knowledge to produce pixel-level but coarse annotations for the images from weak categories. Upon the coarse annotations, the P-Net conducts another knowledge transfer by learning implicit structural fitting patterns between the predicted and manually annotated segmentations in the strong categories to refine the prediction of the weak categories.   

In practice, we notice that segmentation knowledge can be transferred more easily among the categories sharing similar appearances, e.g., from {\ttfamily{bicycle}} to {\ttfamily{motorcycle}}.  Based on this intuitive yet important observation, we devise following learning scheme for the L-Net: we first familiarize the L-Net with the segmentation knowledge learned from strong categories and utilize the  knowledge to predict  class-agnostic segmentation maps of  weak categories with similar appearance but  only image-level labels. Conditioned on  the  segmentation maps by L-Net, a self-diffusion algorithm on the localized  semantic seeds is employed to produce the pixel-level annotations of images from weak categories.

The  P-Net learns to transfer   the verifiable segmentation structural patterns from strong categories and refine  segmentations via adversarial training~\cite{goodfellow2014generative}. Concretely, the P-Net is trained on the strong categories to implicitly learn the fitting patterns between the predicted segmentation map and the raw images, taking ground truth as the adversarial reference. Such knowledge is class-agnostic and well transferable from strong  to weak categories.  P-Net  cannot only tune the prediction to approach the ground truth but also refine details to reduce discrepancies introduced by the inaccurate annotations from L-Net.

We conduct experiments on the PASCAL VOC 2012 dataset, and in case of only $50\%$ ($30\%$) categories with pixel-level annotations, our proposed model achieves $96.5\%$ ($91.4\%$) performance of the fully-supervised baseline. Moreover, we conduct a cross-dataset C3S experiment on transferring the knowledge from completely new categories in MS-COCO to PASCAL VOC 2012 where only image-level labels are available. The proposed model can still retain $89.4\%$ performance of the fully-supervised baseline. Benefiting from the transferable segmentation knowledge from L-Net and the tailored prediction by P-Net, the proposed model can easily produce high-quality pixel-wise masks for a large number of categories, which undoubtedly broadens  image semantic segmentation applications in practice.

\section{Related Work}
To relieve the high demand of pixel-level annotation in semantic segmentation, weakly- and semi-supervised learning approaches have attracted much attention. For weakly-supervised approaches, the image-level label is the simplest way to collect and label. To learn a promising model only with image-level annotations,~\citeauthor{kolesnikov2016seed}~\shortcite{kolesnikov2016seed} defined three loss functions to constrain the model from coarse seeds to fine boundary.~\citeauthor{saleh2016built}~\shortcite{saleh2016built} extracted the activations from higher-level layers as initial segmentation masks.~\citeauthor{kwak2017weakly}~\shortcite{kwak2017weakly} utilized superpixels of the input image as a pooling layout to learn and infer semantic segmentation.~\citeauthor{wei2017object}~\shortcite{wei2017object} progressively mined semantic regions from classification activations to prevent the network from focusing on a small part of an object. Due to the limited information provided by  image-level labels,~\citeauthor{wei2016stc}~\shortcite{wei2016stc} employed saliency maps from extra simple images to provide annotations for learning the semantic segmentation model. Similar to the setting of C3S problem,~\citeauthor{hong2016learning}~\shortcite{hong2016learning} pre-trained an attention model with irrelevant pixel-level annotations for transferring the segmentation knowledge to the weakly labeled targets. Recently,~\citeauthor{hong2017weakly}~\shortcite{hong2017weakly} generated segmentation labels automatically from the web-crawled videos as strong supervision for weakly-supervised semantic segmentation.

\begin{figure*}[t]
	\centering
	\includegraphics[width=0.95\linewidth]{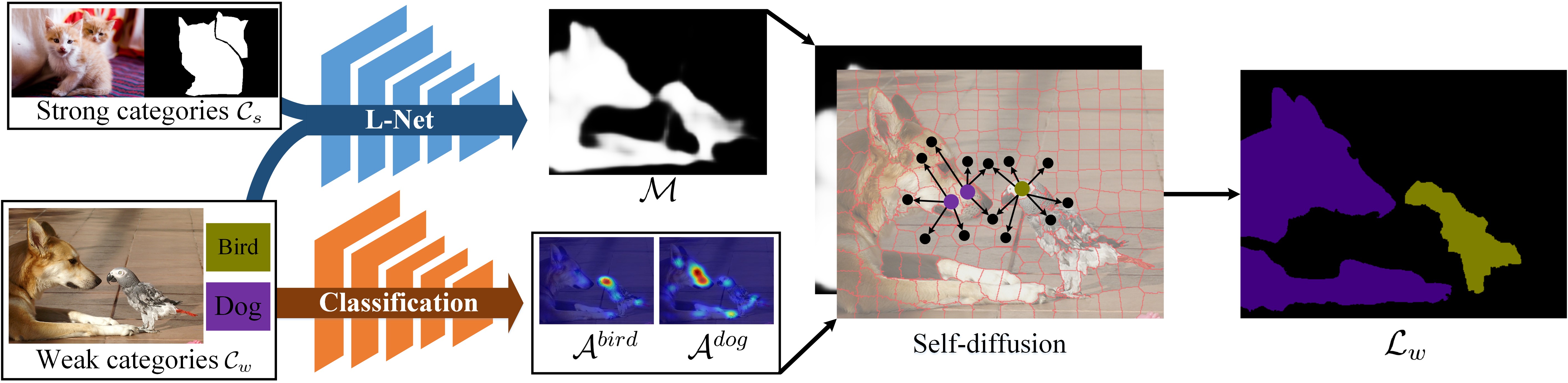}
	\caption{The flowchart of L-Net to produce the pixel-level annotations $\mathcal{L}_w$ for an image from weak categories $\mathcal{C}_w$. L-Net is trained on the images from strong categories with pixel-level annotations (with semantic information removed). Such  more transferable knowledge enables L-Net to produce a class-agnostic segmentation map $\mathcal{M}$ for the image from $\mathcal{C}_w$. Based on the coarse segmentation $\mathcal{M}$, we propagate the class-wise activation maps $\mathcal{A}^{bird}$ and $\mathcal{A}^{dog}$ to generate the final annotations $\mathcal{L}_w$ by a self-diffusion algorithm. Best viewed in color.}
	\label{framework1}
\end{figure*}

Semi-supervised semantic segmentation gives a trade-off between decent performance and labeling efficiency.~\citeauthor{papandreou2015weakly}~\shortcite{papandreou2015weakly} inferred the segmentation model by bundling a fixed proportion of strongly/weakly annotated images in one mini-batch with the expectation maximization methods.~\citeauthor{hong2015decoupled}~\shortcite{hong2015decoupled} separately learned classification and segmentation networks which correspond to different annotations, and transferred the class-specific activations from classification network to segmentation network.~\citeauthor{souly2017semi}~\shortcite{souly2017semi} adopted Generative Adversarial Networks (GANs) to provide extra training samples as a fake class, and the segmentation model acted as a discriminator to classify each pixel to a semantic label or fake label. The above-mentioned semi-supervised approaches are I3S-centric models, meaning they focuses on learning category-specific segmentation knowledge and would fail in case of a new added category. In this work, we attempt to solve the more general and practical C3S problem where annotations at different supervision levels are available across different categories.

\section{Proposed Model}
The proposed model includes two novel components, i.e., the L-Net for learning to produce label maps for weak categories from strong categories and the P-Net for predicting sharp and detailed semantic segmentation. Suppose the weak categories and strong categories are denoted as $\mathcal{C}_w$ and $\mathcal{C}_s$ respectively. The pixel-level annotations for $\mathcal{C}_s$ are denoted as $\mathcal{L}_s$. For the weak categories $\mathcal{C}_w$ provided with only the image-level annotations, the pixel-level annotations $\mathcal{L}_w$ are generated by the L-Net.

\subsection{L-Net: Generating Label Maps for Weak Categories}

To learn the semantic segmentation model, the first step is to produce  pixel-level annotations $\mathcal{L}_w$ for the images from weak categories $\mathcal{C}_w$. In order to provide relatively complete $\mathcal{L}_w$, we introduce the L-Net to learn to perform class-agnostic segmentation as category-agnostic knowledge is easier to learn and transfer among different  categories. The learning process of L-Net is illustrated in Figure~\ref{framework1}. Formally, given training images from the categories $\mathcal{C}_s$ that have pixel-level annotations, the objective for training L-Net (parameterized by $\theta_L$) is defined as follows:
\begin{equation}\label{eq1}
\min_{\theta_L} \sum_{\mathcal{C}_s} \mathcal{J}_b\big(\mathcal{L}_{s}^\prime, \mathcal{O}_L(\mathcal{C}_s;\theta_L)\big),
\end{equation}
where $\mathcal{O}_L(\mathcal{C}_s;\theta_L)$ denotes the output of L-Net, $\mathcal{L}_s^\prime$ is the non-semantic ground truth derived by binarizing $\mathcal{L}_s$ and $\mathcal{J}_b$ denotes the standard element-wise binary cross-entropy loss.

\begin{figure}[t]
	\centering
	\includegraphics[width=1.0\linewidth]{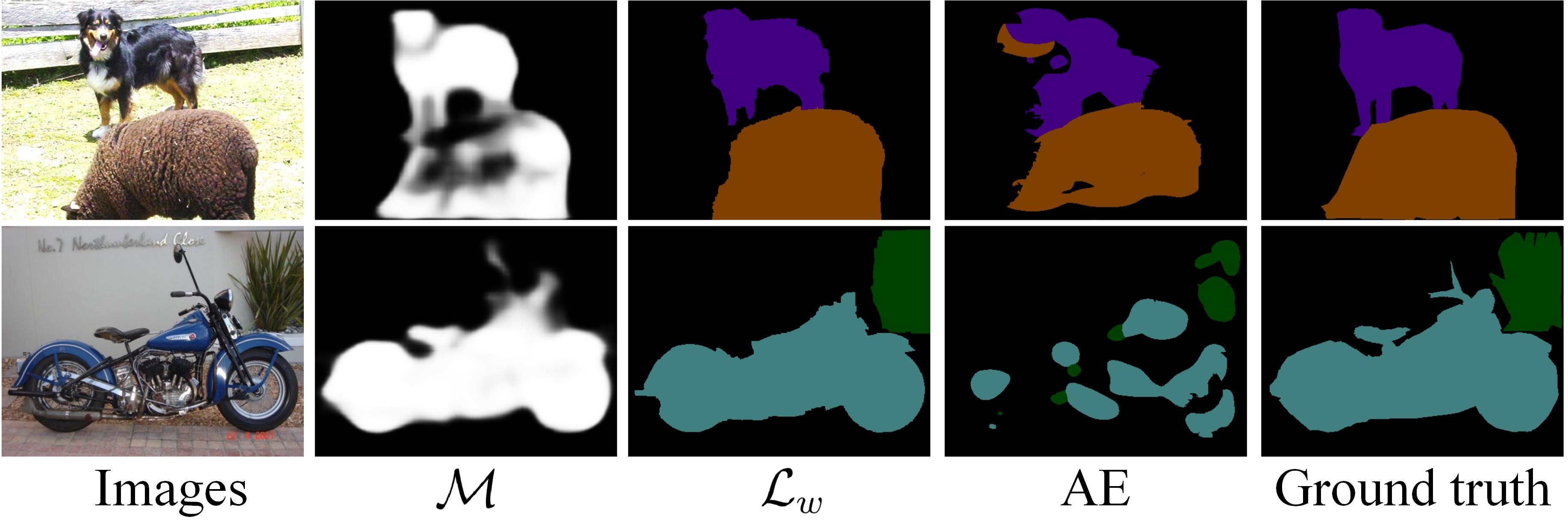}
	\caption{Comparison of generated label maps. $\mathcal{M}$ denotes the class-agnostic segmentation maps while $\mathcal{L}_w$ means the pixel-level annotations generated by the proposed L-Net. AE denotes the Adversarial Erasing approach \cite{wei2017object} to generate the pixel-level annotations for weak categories. One can find that $\mathcal{L}_w$ provides sharp and complete semantic context, even if the $\mathcal{M}$ is noisy. Best viewed in color. \label{fig2}}
\end{figure}

\begin{figure*}[t]
	\centering
	\includegraphics[width=1.0\linewidth]{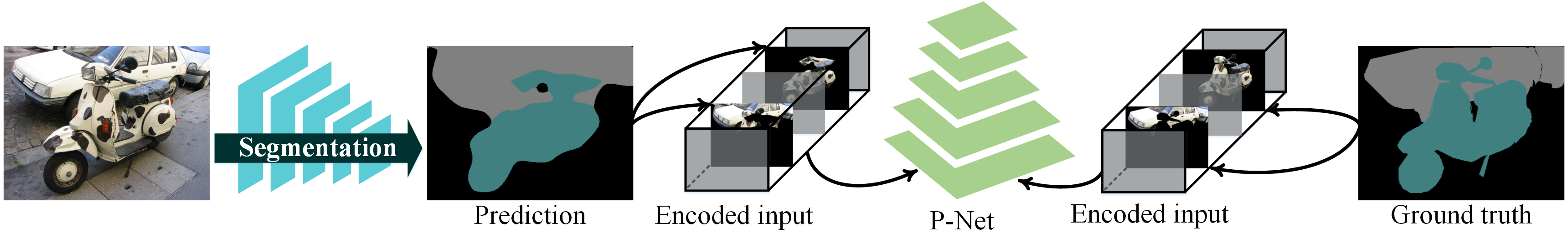}
	\caption{The framework of semantic segmentation with the P-Net. We propose to learn the semantic segmentation model through adversarial training. The input of the P-Net comes from the prediction of the semantic segmentation model and the ground truth, which is encoded by multiplying the training image with the mask of each category. \label{framework2}}
\end{figure*}

The semantic information of $\mathcal{L}_s$ is removed in obtaining $  \mathcal{L}_s^\prime $ in order to learn more transferable knowledge across categories. Such a strategy can fully exploit the object-level information shared among strong categories and benefit segmentation over objects from the weak categories. After training, L-Net is applied to the images of $\mathcal{C}_w$ to produce the class-agnostic segmentation map $\mathcal{M}=\mathcal{O}_L(\mathcal{C}_w;\theta_L)$.

To recover the class-agnostic segmentation map $\mathcal{M}$ to $\mathcal{L}_w$ with rich semantic information, we employ an approach to predict class-discriminative activation by utilizing the image-level annotations available for weak categories. In particular, we employ  a pre-trained image classification network to localize  class-specific activations over the image plane. The bottom panel of Figure~\ref{framework1} visualizes the activation maps $\mathcal{A}^{bird}$ and $\mathcal{A}^{dog}$ produced by a classification network~\cite{zhou2016learning} for two weak categories, {\ttfamily{bird}} and {\ttfamily{dog}} respectively. We take such localization results as reliable seeds for semantic segmentation and diffuse the semantic information originating from these seeds by a Random Walk (RW) based self-diffusion algorithm~\cite{kong2016pattern}. Given an image from $\mathcal{C}_w$, we oversegment it into superpixels $\mathbf{p} = \{ p_1, p_2, \cdots, p_N \}$ which are collectively described by a graph model $G$ where each node corresponds to a particular superpixel. Then, the self-diffusion algorithm is performed on this undirected graph model $G$. Conditioned on $\mathcal{M}$, the objective function of the self-diffusion process for a specific category  $\mathcal{A}^c$ is defined as
\begin{equation}\label{eq2}
\min_{\mathbf{q}} \frac{1}{2}\sum_{i,j}z_{ij}(q_i-q_j)^2,
\end{equation}
where $\mathbf{q} = [q_1, q_2, \cdots, q_N]$ denotes the label vector of all superpixels $\mathbf{p}$. If $p_i \in \mathcal{A}^c$, $q_i$ is fixed to 1, and otherwise it takes an initial value of 0. $z_{ij}=\exp(-\|\mathcal{F}(p_i)-\mathcal{F}(p_j)\|/2\sigma^2)$ denotes the Gaussian distance between two adjacent superpixels. $\mathcal{F}(p_i)\in\mathbb{R}^4$ denotes the mean feature of superpixel $p_i$ in the normalized CIELAB color space and the segmentation map $\mathcal{M}$. 

Eqn.~(\ref{eq2}) formulates the conventional RW algorithm that strengthens label consistency of nodes with large affinity. Considering there are some examples hard to be transferred within the L-Net, e.g., the {\ttfamily{plant}} in the second example of Figure~\ref{fig2}, we impose no extra constraint on the segmentation map $\mathcal{M}$ in Eqn.~(\ref{eq2}). When the L-Net cannot segment all objects out, the high confidence class activation maps can still reveal and propagate segmentation information well. If there are more than two weak categories in the image to segment, we assign the category label $c$ to the superpixel $p_i$ with a larger $q_i$. As shown in Figure~\ref{fig2}, we compare the generated label map by self-diffusion with the State-Of-The-Art (SOTA) Adversarial Erasing (AE) approach~\cite{wei2017object}. Observing the generated pixel-level annotations, one can find that $\mathcal{L}_w$ provides  semantic context at a satisfactory level, even if the segmentation map $\mathcal{M}$ is noisy.

\subsection{P-Net: Semantic Segmentation with Adversarial Learning}
Once L-Net generates the coarse pixel-annotations of weak categories, the semantic segmentation model can be trained upon such annotations. However, to get sharper and more accurate segmentation results, we introduce the P-Net component that learns to refine the semantic segmentation with adversarial training \cite{goodfellow2014generative}, as shown in Figure~\ref{framework2}. The generator within the adversarial learning framework is the semantic segmentation model in the left of Figure~\ref{framework2} which tries to predict label maps to match the joint data distribution of the ground truth and input images. The discriminator called P-Net acts to distinguish the input drawn from the generator or from the ground truth. On the one hand, the adversarial training forces the prediction of the semantic segmentation model to be  as close as possible to the ground truth. On the other hand, the adversarial training  learns to capture and utilize the implicit fitting patterns between the prediction and the ground truth which can be transferred to the weak categories.

Formally, for a given training sample $I$ and its corresponding label map $\mathcal{L}_I$, we define the objective of adversarial training as follows:
\small
\begin{equation}\label{eq3}
\begin{aligned}
&\min_{\theta_S}\max_{\theta_P} \sum_I \mathcal{J}_m\big(\mathcal{L}_I, \mathcal{O}_S(I;\theta_S)\big) - \\
&\lambda\Big[\mathcal{J}_b\big(1, \mathcal{O}_P(\mathcal{L}_I;\theta_P)\big) + \mathcal{J}_b\big(0, \mathcal{O}_P(\mathcal{O}_S(I;\theta_S);\theta_P)\big)\Big],
\end{aligned}
\end{equation}
\normalsize
where $\theta_S$ and $\theta_P$ denote the parameters of the semantic segmentation model and the P-Net respectively. $\mathcal{J}_m$ and $\mathcal{J}_b$ denote the multi-class and binary cross-entropy loss respectively. $\mathcal{O}_S$ and $\mathcal{O}_P$ denote the output of the semantic segmentation model and the P-Net respectively. We use 1 and 0 to denote the label of P-Net when its input comes from the ground truth $\mathcal{L}_I$ and the prediction $\mathcal{O}_S(I;\theta_S)$  respectively.

For training the semantic segmentation model, we minimize the loss in Eqn.~\eqref{eq3} w.r.t. $\theta_S$:
\small
\begin{equation}\label{eq4}
\min_{\theta_S} \sum_I \mathcal{J}_m\big(\mathcal{L}_I, \mathcal{O}_S(I;\theta_S)\big) + \lambda\mathcal{J}_b\big(1, \mathcal{O}_P(\mathcal{O}_S(I;\theta_S);\theta_P)\big),
\end{equation}
\normalsize
where the term $\lambda\mathcal{J}_b\big(1, \mathcal{O}_P(\mathcal{O}_S(I;\theta_S);\theta_P)\big)$ replaces the term $-\lambda\mathcal{J}_b\big(0, \mathcal{O}_P(\mathcal{O}_S(I;\theta_S);\theta_P)\big)$ in Eqn.~\eqref{eq3}. The first term in Eqn.~\eqref{eq4} encourages the prediction of semantic segmentation to be consistent with the ground truth at each position while the second term penalizes the unfitting structure between the prediction and the ground truth.

For training the P-Net, we minimize the loss in Eqn.~\eqref{eq3} w.r.t. $\theta_P$:
\small
\begin{equation}\label{eq5}
\min_{\theta_P} \sum_I \Big[\mathcal{J}_b\big(1, \mathcal{O}_P(\mathcal{L}_I;\theta_P)\big) + \mathcal{J}_b\big(0, \mathcal{O}_P(\mathcal{O}_S(I;\theta_S);\theta_P)\big)\Big].
\end{equation}
\normalsize

Inspired by~\citeauthor{luc2016semantic}~\shortcite{luc2016semantic}, we do not directly input the probability maps predicted by the semantic segmentation network to P-Net. Instead, as shown in Figure~\ref{framework2}, we encode the input of P-Net by multiplying the training image $I$ with the predicted segmentation mask $\mathcal{O}_S(I;\theta_S)$ or the ground truth mask  $\mathcal{L}_I$. This encoding makes the P-Net observe different objects and does not emphasize too much on the semantic label, which facilitates  knowledge transfer across categories. Considering the unreliable label maps generated by the L-Net, directly training the whole network in Figure~\ref{framework2} could lead to poor performance of the P-Net, because the generated label maps may fall in conflict with the ground truth from strong categories. Therefore, we first pre-train the P-Net with the strong categories to encourage the P-Net to learn the real high-order fitting patterns and then fine-tune the whole training set. Experiments in the following section prove that it is indeed helpful to improve performance on weak categories.

\section{Experiments}\label{sec4}
\begin{table}[t]
	\caption{Layer configuration of P-Net}\label{tab_dnet}
	\footnotesize
	\centering
	\begin{tabular}{cccc}
		\toprule
		Layer & Channels & Kernel & Activation \\
		\midrule
		conv1 & $\begin{array}{l}16 \\ 32  \end{array} $ & $\begin{array}{l}\textsc{3$\times$3} \\ \textsc{3$\times$3}  \end{array} $ & ReLU \vspace{0.1cm} \\ 
		pool1 & - &  2$\times$ 2, stride 2 & - \\
		\midrule
		conv2 & $\begin{array}{l}64 \\ 64  \end{array} $ & $\begin{array}{l}\textsc{3$\times$3} \\ \textsc{3$\times$3}  \end{array} $ & ReLU \vspace{0.1cm} \\
		pool2 & - & 2$\times$ 2, stride 2 & - \\
		\midrule
		conv3 & $\begin{array}{l}128 \\ 128  \end{array} $ & $\begin{array}{l}\textsc{3$\times$3} \\ \textsc{3$\times$3}  \end{array} $  & ReLU \vspace{0.1cm} \\
		pool3 & & 2$\times$ 2, stride 2 & - \\
		\midrule
		fc4 & 256-d & - & tanh \\
		fc5 & 512-d & - & tanh \\
		fc6 & 1-d & - & sigmoid \\
		\bottomrule
	\end{tabular}
\end{table} 

\subsection{Implementation Details}
\subsubsection{Datasets} 
We evaluate the performance of the proposed model on the PASCAL VOC 2012 benchmark~\cite{2010-pascal} which contains one background category and 20 object categories. The training set contains 10,582 images with pixel-level annotations, which is extended by~\citeauthor{hariharan2011semantic}~\shortcite{hariharan2011semantic}. We evaluate the performance in terms of mean Intersection over Union (mIoU) on other two subsets, i.e., validation and test, including 1,449 and 1,456 images respectively. According to the appearance similarity, we divide the 20 object categories into two super-categories, i.e., strong categories and weak categories, to guarantee each super-category contains similar categories. We provide four split-sets of the training images. Split-set 1 consists of 10 strong categories and 10 weak categories while split-set 2 is reversed based on split-set 1. Split-set 3 is a harder case which contains 6 strong categories and 14 weak categories. Similar to the setting in ~\citeauthor{hong2016learning}~\shortcite{hong2016learning}, split-set 4 only provides  image-level annotations for all 20 categories in PASCAL VOC 2012 while the strong categories are derived from MS-COCO~\cite{lin2014microsoft}. The training images containing PASCAL VOC 2012 categories are removed from MS-COCO and the remaining 16,241 images from 60 exclusive categories are employed as strong categories.

\begin{table*}[!t]\setlength{\tabcolsep}{1.8pt}
	\footnotesize
	\centering
	\caption{Performance on PASCAL VOC 2012 validation set. The number in {\setlength{\fboxsep}{1.5pt}\colorbox{mygray2}{gray block}} represents the performance of categories with \emph{pixel-level} annotations.} 
	\label{tab1}
	\begin{tabular}{lccccccccccccccccccccc|c}
		\toprule
		& bkg & aero & bike  & bird  & boat  & bottle & bus   & car   & cat   & chair & cow   & table & dog   & horse & mbk & prsn & plnt & sheep & sofa  & train & tv &  mIoU  \\
		\midrule
		\multicolumn{22}{l}{(a) Fully supervised baseline}\\
		DeepLab   & 90.2  & 73.2  & 31.0  & 73.7  & 58.6  & 63.5  & 81.4  & 74.5  & 76.9  & 28.4  & 62.7  & 49.9  & 69.4  & 
		61.5  & 67.1  & 76.1  & 46.2  & 67.8  & 42.0  & 74.1  & 52.6  & 62.9 \\
		\midrule
		\multicolumn{22}{l}{(b) Split-set 1}\\
		WSSL$^\dagger$ & 83.3 & 31.1 & \ttt{31.2}& 29.1 & \ttt{59.9} & 31.9 & \ttt{80.6} & \ttt{74.3} & \ttt{78.2} & \ttt{29.7} & \ttt{61.6} & \ttt{43.7} & 39.6 & 38.6 & 29.5 & \ttt{74.8} & 21.5 & 33.4 & 19.1 & 45.5 & \tt{48.4} & 46.9 \\
		AE & 85.8  & 50.5  & \ttt{31.0}  & 44.6  & \ttt{56.8}  & 40.7 & \ttt{78.7} & \ttt{74.1}  & \ttt{75.1}  & \ttt{27.6}  & \ttt{61.2}  & \ttt{50.0}  & 54.1  & 42.8  & 51.1  & \ttt{75.8}  & 28.2  & 53.3  & 26.9  & 46.2 & \tt{53.1}  & 52.7 \\
		L-Net & 89.6  & 74.0  & \ttt{30.8}  & 66.8  & \ttt{58.7}  & 43.1  & \ttt{80.8}  & \ttt{75.6}  & \ttt{76.3}  & \ttt{28.3}  & \ttt{61.7}  & \ttt{48.8}  & 66.5  & 60.7  & 68.2  & \ttt{76.0}  & 30.4  & 68.8  & 27.4  & 65.8  & \tt{53.1}  & 59.6 \\
		P-Net & 90.0  & 74.0  & \ttt{30.9}  & 64.3  & \ttt{59.6}  & 43.4  & \ttt{82.8}  & \ttt{76.7}  & \ttt{77.7}  & \ttt{27.8}  & \ttt{66.0}  & \ttt{51.2}  & 68.4  & 63.7  & 68.5  & \ttt{76.7}  & 33.4  & 71.6  & 28.4  & 64.3  & \tt{55.0}  & 60.7 \\
		\midrule
		\multicolumn{22}{l}{(c) Split-set 2}\\
		WSSL$^\dagger$ & 82.1 & \ttt{77.3} & 17.4 & \ttt{73.5} & 29.1 & \ttt{63.1} & 45.3 & 40.2 & 43.2 & 16.5 & 35.4 & 27.3 & \ttt{69.0} & \ttt{56.3} & \ttt{61.2} & 27.4 & \ttt{45.1} & \ttt{69.4} & \ttt{28.9} & \ttt{73.9} & 33.1 & 48.3\\
		AE & 84.2  & \ttt{72.1}  & 22.9  & \ttt{71.8}  & 32.6  & \ttt{61.0}  & 63.6  & 30.0  & 59.5  & 16.0  & 43.2  & 22.7  & \ttt{67.5}  & \ttt{58.7}  & \ttt{65.4}  & 53.0  & \ttt{45.4}  & \ttt{66.9}  & \ttt{37.7}  & \ttt{70.9}  & 39.9  & 51.7 \\
		L-Net & 87.1  & \ttt{72.8}  & 30.7  & \ttt{71.7}  & 50.6  & \ttt{62.4}  & 76.3  & 71.3  & 73.2  & 17.5  & 59.1  & 15.4  & \ttt{68.3}  & \ttt{60.9}  & \ttt{65.5}  & 50.6  & \ttt{43.5}  & \ttt{67.9}  & \ttt{39.5}  & \ttt{71.4}  & 45.7  & 57.2 \\
		P-Net & 87.7  & \ttt{74.4}  & 31.1  & \ttt{72.6}  & 53.9  & \ttt{62.7}  & 77.1  & 73.0  & 73.9  & 17.5  & 61.8  & 16.6  & \ttt{70.3}  & \ttt{62.2}  & \ttt{66.8}  & 51.5  & \ttt{45.0}  & \ttt{69.3}  & \ttt{40.0}  & \ttt{72.8}  & 49.0  & 58.5 \\
		\midrule
		\multicolumn{22}{l}{(d) Split-set 3}\\
		WSSL$^\dagger$ & 82.4 & 31.0 & 15.2 & 30.1 & 26.2 & 33.2 & 44.7 & 42.2 & 45.1 & \ttt{23.6} & \ttt{64.4} & 25.6 & \ttt{70.3} & 38.4 & \ttt{54.4} & \ttt{76.5} & 25.1 & 33.6 & 20.6 & \ttt{75.9} & 33.8 & 42.5    \\
		AE & 85.1  & 50.0  & 23.4  & 46.6  & 31.8  & 39.7  & 62.8  & 30.7  & 59.4  & \ttt{28.8}  & \ttt{61.7}  & 25.7  & \ttt{67.6}  & 42.6  & \ttt{65.0}  & \ttt{76.6}  & 28.8  & 52.6  & 28.2  & \ttt{72.3}  & 39.0  & 48.5 \\
		L-Net & 89.1  & 58.3  & 33.5  & 71.1  & 34.8  & 42.0  & 75.4  & 67.7  & 74.9  & \ttt{27.1}  & \ttt{62.2}  & 24.5  & \ttt{70.5}  & 61.1  & \ttt{67.5}  & \ttt{76.4}  & 31.2  & 68.4  & 25.5  & \ttt{74.2}  & 47.5  & 56.4 \\
		P-Net & 89.3  & 57.2  & 34.5  & 71.2  & 38.6  & 44.5  & 77.7  & 67.5  & 76.2  & \ttt{24.3}  & \ttt{64.5}  & 25.8  & \ttt{72.9}  & 63.8  & \ttt{69.5}  & \ttt{77.3}  & 32.1  & 71.8  & 27.4  & \ttt{77.9}  & 44.1  & 57.5 \\
		\midrule
		\multicolumn{22}{l}{(e) Split-set 4}\\
		TransferNet & 85.3 & 68.5 & 26.4 & 69.8 & 36.7 & 49.1 & 68.4 & 55.8 & 77.3 & 6.2 & 75.2 & 14.3 & 69.8 & 71.5 & 61.1 & 31.9 & 25.5 & 74.6 & 33.8 & 49.6 & 43.7 & 52.1 \\
		L-Net & 86.5 & 70.9 & 26.3 & 70.1 & 46.3 & 55.3 & 73.9 & 67.6 & 72.9 & 20.6 & 61.4 & 21.2 & 66.1 & 60.5 & 63.2 & 55.2 & 32.7 & 66.3 & 34.2 & 64.4 & 44.9 & 55.3 \\
		P-Net & 87.1 & 71.2 & 25.1 & 69.9 & 48.5 & 56.2 & 75.7 & 67.1 & 75.0 & 19.2 & 63.8 & 22.3 & 67.0 & 64.2 & 62.9 & 55.1 & 35.2 & 69.6 & 34.4 & 67.2 & 43.1 & 56.2  \\
		\bottomrule
	\end{tabular}
\end{table*}

\subsubsection{Network Architecture}
In this paper, we focus on transfer learning across various categories with different types of annotations. Therefore, extensive engineering on the segmentation network architecture is out of the scope of this work. We adopt the popular architecture of DeepLab-LargeFOV~\cite{chen2014semantic} as the backbone network for the L-Net in Figure~\ref{framework1} and the semantic segmentation network in Figure~\ref{framework2}. DeepLab-LargeFOV is initialized by the weights of VGG-16 model~\cite{simonyan2014very} which is pre-trained on the ImageNet. The L-Net differs from the semantic segmentation network in the loss function as shown in Eqn.~(\ref{eq1}) and Eqn.~(\ref{eq5}). The classification model in Figure~\ref{framework1} that provides category-specific activation maps is identical with the VGG-16 based CAM model~\cite{zhou2016learning} and is fine-tuned on PASCAL VOC 2012 dataset with image-level labels. The P-Net in Figure~\ref{framework2} consists of six $3\times3$ convolutional layers and three fully connected layers. Details on layer configuration are provided in Tabel~\ref{tab_dnet}.

\subsubsection{Training} 
For the training of L-Net, we convert the semantic label maps from strong categories to a binary mask. We take a mini-batch size of 30, in which patches of 321$\times$321 pixels are randomly cropped from images. We totally perform 30 epochs for training the L-Net with an initial learning rate of 5e-8. Momentum and weight decay are set to 0.9 and 0.0005 respectively. We train the semantic segmentation model in the same setting as DeepLab-LargeFOV. When the semantic segmentation model is trained, we fine-tune it with scratched P-Net and set the learning rate of the semantic segmentation model and the P-Net to 1e-5 and 1e-3 respectively. All the experiments are performed on NVIDIA TITAN X PASCAL GPU with 12G memory. 

\subsection{Comparison with Baselines}\label{sec4.2}
We evaluate various models on the PASCAL VOC 2012 validation set with four different strong/weak category splits. The results are summarized in Table~\ref{tab1}. In particular, we compare the proposed model with following four baselines.
\begin{enumerate}
	\item We use the fully-supervised DeepLab-LargeFOV~\cite{chen2014semantic} to gain performance  upper bound for the compared weakly-/semi-supervised segmentation methods. 
	\item We also compare with an I3S-centric model, WSSL~\cite{papandreou2015weakly}. It has the same segmentation network, i.e., DeepLab-LargeFOV, as our proposed model and  is directly applied for the C3S problem introduced in this work. Following  the practice in ~\citeauthor{papandreou2015weakly}~\shortcite{papandreou2015weakly}, the semantic segmentation for weak categories is inferred via the adaptive EM algorithm.
	\item We adopt a SOTA weakly-supervised approach, i.e., AE~\cite{wei2017object}, as the third  baseline,  aiming to thoroughly compare model's capability of predicting pixel-level annotations for weak categories. During evaluation of AE, we apply AE to generate  semantic label maps of the weak categories at first. Then we train and evaluate the DeepLab-LargeFOV model using the AE generated label maps (for weak categories) and the provided ground truths (for strong categories).
	\item For the fourth split-set, we also compare with the TransferNet~\cite{hong2016learning}. We evaluate its performance using a stronger segmentation model DeconvNet~\cite{noh2015learning}, under the same setting as our proposed model.
\end{enumerate} 

In Table~\ref{tab1}, the numbers in {\setlength{\fboxsep}{1.0pt}\colorbox{mygray2}{gray block}} represent the segmentation performance of strong categories. The ``L-Net''  denotes the results obtained  by the semantic segmentation model  trained on both the strong categories and the  label maps generated by L-Net. The ``P-Net'' denotes the final results by applying the P-Net to refine the semantic segmentation results. From the results, one can make following observations.  The results of WSSL$^\dagger$ in Table~\ref{tab1} demonstrate that the I3S-centric WSSL~\cite{papandreou2015weakly} performs poorly for the weak categories because it is incapable of transferring  knowledge  across categories.  For the newly introduced category, WSSL performs not so well without additional pixel-level annotations. The proposed L-Net outperforms SOTA AE~\cite{wei2017object} by $13.1\%$, $10.6\%$ and $16.3\%$ on the first three split-sets respectively, confirming   effectiveness of the L-Net on predicting high-quality  label maps. For some weak categories (e.g., the category {\ttfamily{motorbike}} in split-set 1 and the category {\ttfamily{horse}} in split-set 3), L-Net even performs slightly better than the fully-supervised model. We attribute this surprising superiority to the useful  knowledge transferred  across categories with similar appearance. For  split-set 4, L-Net  improves over TransferNet~\cite{hong2016learning} by $6.1\%$ under the same setting, proving  the transferable segmentation knowledge in L-Net  is more appropriate than the attention-based mechanism in TransferNet~\cite{hong2016learning}. 

\begin{figure*}[t]
	\centering
	\includegraphics[width=0.93\linewidth]{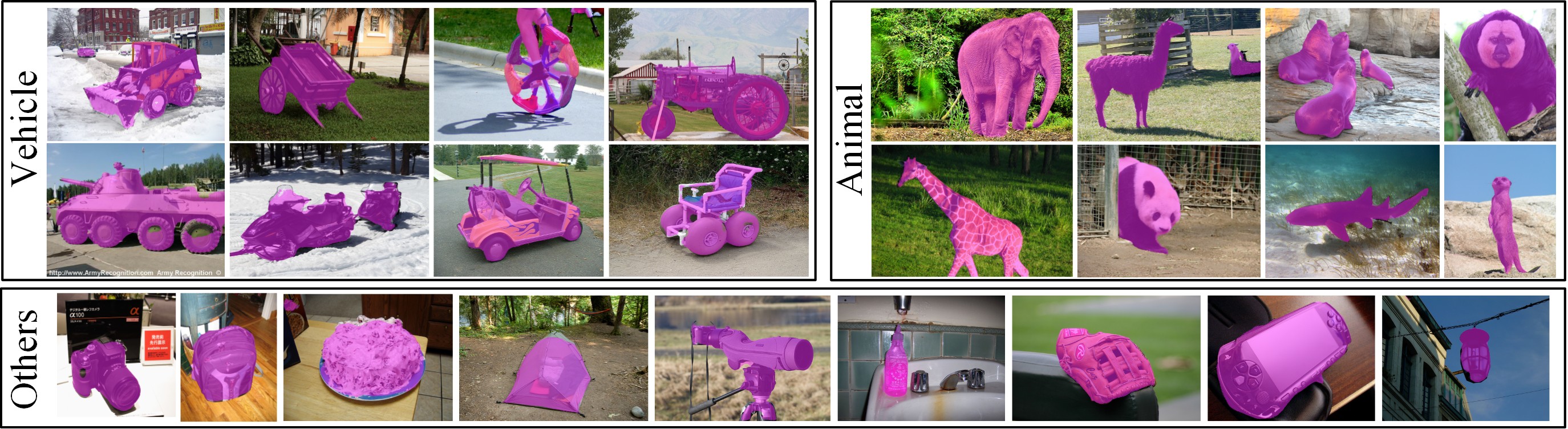}
	\caption{Segmentation results on unseen categories from ImageNet~\cite{russakovsky2015imagenet}. All the results are produced by the L-Net which is trained on the split-set 1. Best viewed in color.}
	\label{fig4}
\end{figure*}

As shown in Table~\ref{tab1}, employing  adversarial training over a semantic segmentation model  further improves the results by $1.9\%$. The P-Net pre-trained by the strong categories can learn the implicit fitting patterns between the prediction and the ``real'' pixel-level annotations. The learned suitable knowledge can be transferred to the weak categories and alleviate high-level disparities in the prediction of images from weak categories. We observe that pre-training on the strong categories is useful for stabilizing  training process  of P-Net. This is because some pixel-level annotations in weak categories are not  reliable and may contaminate  P-Net. If we directly train P-Net with the whole training set (consisting of provided pixel-level annotations and predicted ones from L-Net), we find the improvement brought by P-Net on split-set 1 is only 0.5$\%$\textemdash on the other three split-sets the performance may even drop. Overall, the proposed model provides a very promising solution for segmenting the categories without pixel-level annotations and approaches the performance of the fully-supervised baseline. 

\begin{table}[h!]
	\centering
	\small
	\caption{Comparison with weakly- and semi-supervised semantic segmentation models on PASCAL VOC 2012 test set.}\label{tab_com} 
	\begin{tabular}{lcc}
		\toprule
		Methods &  $\#$Training Set & mIoU \\
		\midrule
		\multicolumn{3}{l}{(a) Weakly-supervised methods}\\
		DCSM~\shortcite{shimoda2016distinct} & 10k  & 45.1 \\
		BFBP~\shortcite{saleh2016built} & 10k  & 48.0 \\
		STC~\shortcite{wei2016stc} & 50k  & 51.2 \\
		SEC~\shortcite{kolesnikov2016seed} & 10k & 51.7 \\
		FCL~\shortcite{roy2017combining} & 10k & 53.7 \\
		AE~\shortcite{wei2017object} & 10k  & 55.7 \\
		\citeauthor{hong2017weakly}~\shortcite{hong2017weakly} & 970k  & 58.7 \\ 
		\midrule
		\multicolumn{3}{l}{(b) In-category semi-supervised methods}\\
		WSSL~\shortcite{papandreou2015weakly} & 10k & 66.2 \\
		DecoupledNet~\shortcite{hong2015decoupled} & 10k & 62.5  \\
		\midrule
		\multicolumn{3}{l}{(c) Cross-category semi-supervised methods}\\
		Ours (Split-set 1) & 10k & 64.6\\
		Ours (Split-set 2) & 10k & 61.9\\
		Ours (Split-set 3) & 10k & 59.5\\
		Ours (Split-set 4) & 27k & 58.0\\
		TransferNet (Split-set 4) & 27k  & 51.2 \\
		\bottomrule
	\end{tabular}
\end{table} 

\subsection{Comparison with State-of-the-arts}
We further compare our proposed model with several SOTA weakly- and semi-supervised semantic segmentation models, provided  with different levels of annotations. Table~\ref{tab_com} presents relevant results on PASCAL VOC 2012 test set. Among the compared models, Ours (Split-set 4), TransferNet~\cite{hong2016learning}, STC~\cite{wei2016stc} and ~\citeauthor{hong2017weakly}~\shortcite{hong2017weakly} employ extra data (16k, 16k, 40k and 960k) for segmentation while the other methods are based on the 10k training samples of PASCAL VOC 2012. The pixel-level annotations in DecoupledNet~\cite{hong2015decoupled} are provided for 500 images while the number in WSSL is 1,464. For fair comparison, we apply post-processing over the  results from  P-Net  with CRF~\cite{koltun2011efficient}. Compared with the latest weakly-supervised method~\cite{hong2017weakly}, Ours (Split-set 4) performs competitively well as the model trained using 960k extra images in~\citeauthor{hong2017weakly}~\shortcite{hong2017weakly}. However, the proposed model only uses the  image-level annotations of PASCAL VOC 2012 and 16k irrelevant pixel-level annotations.

For the semi-supervised semantic segmentation, even though as few as $1/2$ categories have pixel-level annotations in Ours (Split-set 1), the performance of proposed model only degrades by $2.5\%$ compared with I3S-centric WSSL. Actually, based on the results in Table~\ref{tab1}, the I3S-centric approaches~\cite{papandreou2015weakly,hong2015decoupled} cannot handle the C3S problem well and fail to generalize the weak categories. Such deficiency will restrain their application to the newly introduced categories. 
Compared with the attention-based TransferNet~\cite{hong2016learning}, the proposed model (Split-set 4) is advantageous. It introduces two complementary transferable components on segmentation knowledge and can provide superior semantic segmentation results as shown in Table~\ref{tab_com}. 

\begin{figure}[!t]
	\centering
	\includegraphics[width=0.99\linewidth]{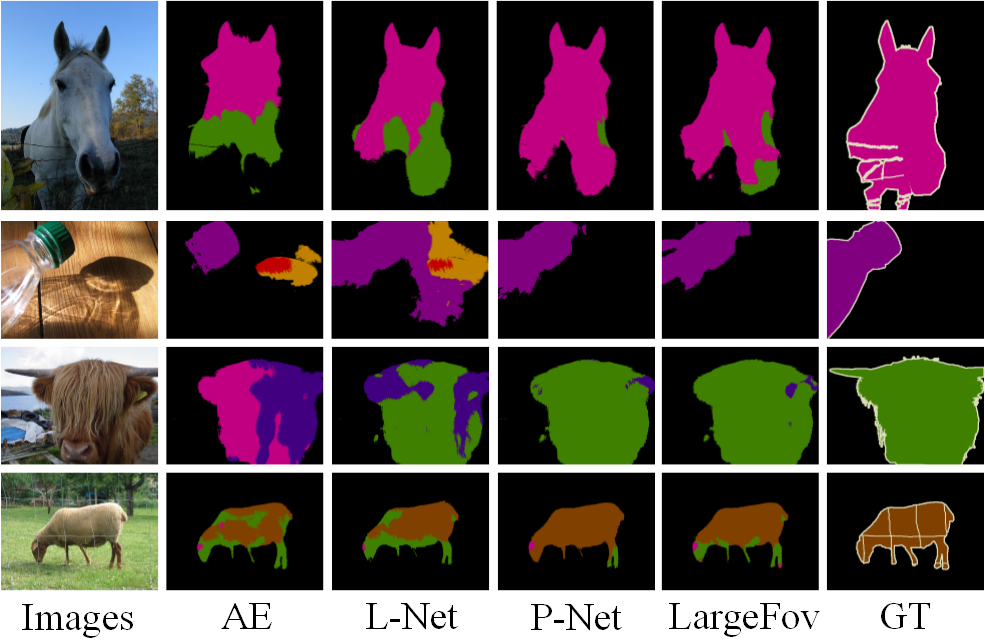}
	\caption{Visual comparison of semantic segmentation results. AE denotes the segmentation results by the weakly-supervised baseline~\cite{wei2017object} while LargeFov denotes the results by the DeepLab-LargeFov. P-Net denotes the refined results of L-Net by adversarial training. The first two examples come from the weak categories of split-set 1 while the last two examples come from the weak categories of split-set 2 and split-set 3 respectively. Best viewed in color.
	}
	\label{fig3}
\end{figure}

\subsection{Running Time}
In this paper, training the L-Net with 3,000 images for 30 epochs takes about 3 hours while the inference of self-diffusion algorithm takes only 1 second for an input image. Training the P-Net with 10,000 images for 30 epochs takes about 12 hours. The total training time of the proposed method is about 17 hours (the training time of L-Net and P-Net plus self-diffusion inference on 7,000 weakly labeled images). The time cost is comparable with WSSL~\cite{papandreou2015weakly} which takes about 10 hours with the same setting. For testing, the proposed model has the same computational complexity as WSSL and it takes about 0.2 second to process a 300$\times$400 image.

\subsection{Qualitative Results}
To verify the effectiveness of the learned L-Net, we apply L-Net on the unseen categories from ImageNet as shown in Figure~\ref{fig4}. All the results in Figure~\ref{fig4} are produced by the L-Net  trained on the split-set 1 of PASCAL VOC 2012. One can find that the L-Net generalizes well on those unseen categories and provides sharp and complete segmentation masks. The  L-Net generalizes well and provides practical solution for transferring a segmentation model from familiar objects to unseen ones. In Figure~\ref{fig3}, we provide visual comparisons of the semantic segmentation results by AE~\cite{wei2017object}, L-Net, P-Net and DeepLab-LargeFov~\cite{chen2014semantic}. The first two examples come from the weak categories of split-set 1 and the last two examples come from the weak categories of split-set 2 and split-set 3 respectively. From the results of P-Net, one can observe that the adversarial training can clean the noisy regions of L-Net and maintain consistency with the ground truth. 

\section{Conclusion}
In this paper we tackle a more general problem in semi-supervised semantic segmentation where the strong categories and weak categories do not have overlap. We propose a novel transferable semi-supervised semantic segmentation model which contains two networks capable of learning and transferring segmentation knowledge, i.e., L-Net and P-Net. The L-Net generates the label maps of weak categories while the P-Net further refines the transferred knowledge by correcting high-level discrepancies between the prediction and ground truth. Benefited from the cross-category transferring, the proposed model provides superior performance over SOTA weakly-supervised approaches on the newly added category. Though only a small fraction of categories are with pixel-level annotations, the proposed model can still achieve $90\%$ performance of the fully-supervised baseline. It enhances the applicability and scalability of semantic segmentation models in real applications.

\subsubsection{Acknowledgments}
This work was supported in part by the National Natural Science Foundation of China under Grant 61403403, China Postdoctoral Science Foundation under Grant 2015M52707, and the China Scholarship Council under Grant
201603170287. The work of Yunchao Wei was partially supported by IBM-ILLINOIS Center
for Cognitive Computing Systems Research (C3SR)---a research collaboration as part of the IBM AI Horizons Network. The work of Jiashi Feng was partially supported by NUS startup R-263-000-C08-133, MOE Tier-I R-263-000-C21-112 and IDS R-263-000-C67-646.

\bibliography{egbib}

\begin{thebibliography}{}

\bibitem[\protect\citeauthoryear{Chen \bgroup et al\mbox.\egroup
  }{2015}]{chen2014semantic}
Chen, L.-C.; Papandreou, G.; Kokkinos, I.; Murphy, K.; and Yuille, A.~L.
\newblock 2015.
\newblock Semantic image segmentation with deep convolutional nets and fully
  connected crfs.
\newblock In {\em {ICLR}}.

\bibitem[\protect\citeauthoryear{Everingham \bgroup et al\mbox.\egroup
  }{2014}]{2010-pascal}
Everingham, M.; Eslami, S.~A.; Van~Gool, L.; Williams, C.~K.; Winn, J.; and
  Zisserman, A.
\newblock 2014.
\newblock The pascal visual object classes challenge: A retrospective.
\newblock {\em {IJCV}} 111(1):98--136.

\bibitem[\protect\citeauthoryear{Goodfellow \bgroup et al\mbox.\egroup
  }{2014}]{goodfellow2014generative}
Goodfellow, I.; Pouget-Abadie, J.; Mirza, M.; Xu, B.; Warde-Farley, D.; Ozair,
  S.; Courville, A.; and Bengio, Y.
\newblock 2014.
\newblock Generative adversarial nets.
\newblock In {\em {NIPS}}.

\bibitem[\protect\citeauthoryear{Hariharan \bgroup et al\mbox.\egroup
  }{2011}]{hariharan2011semantic}
Hariharan, B.; Arbel\"aez, P.; Bourdev, L.; Maji, S.; and Malik, J.
\newblock 2011.
\newblock Semantic contours from inverse detectors.
\newblock In {\em {ICCV}}.

\bibitem[\protect\citeauthoryear{Hong \bgroup et al\mbox.\egroup
  }{2016}]{hong2016learning}
Hong, S.; Oh, J.; Lee, H.; and Han, B.
\newblock 2016.
\newblock Learning transferrable knowledge for semantic segmentation with deep
  convolutional neural network.
\newblock In {\em {CVPR}}.

\bibitem[\protect\citeauthoryear{Hong \bgroup et al\mbox.\egroup
  }{2017}]{hong2017weakly}
Hong, S.; Yeo, D.; Kwak, S.; Lee, H.; and Han, B.
\newblock 2017.
\newblock Weakly supervised semantic segmentation using web-crawled videos.
\newblock In {\em {CVPR}}.

\bibitem[\protect\citeauthoryear{Hong, Noh, and Han}{2015}]{hong2015decoupled}
Hong, S.; Noh, H.; and Han, B.
\newblock 2015.
\newblock Decoupled deep neural network for semi-supervised semantic
  segmentation.
\newblock In {\em {NIPS}}.

\bibitem[\protect\citeauthoryear{Kolesnikov and
  Lampert}{2016}]{kolesnikov2016seed}
Kolesnikov, A., and Lampert, C.~H.
\newblock 2016.
\newblock Seed, expand and constrain: Three principles for weakly-supervised
  image segmentation.
\newblock In {\em {ECCV}}.

\bibitem[\protect\citeauthoryear{Kong \bgroup et al\mbox.\egroup
  }{2016}]{kong2016pattern}
Kong, Y.; Wang, L.; Liu, X.; Lu, H.; and Ruan, X.
\newblock 2016.
\newblock Pattern mining saliency.
\newblock In {\em {ECCV}}.

\bibitem[\protect\citeauthoryear{Kr\"ahenb\"uhl and
  Koltun}{2011}]{koltun2011efficient}
Kr\"ahenb\"uhl, P., and Koltun, V.
\newblock 2011.
\newblock Efficient inference in fully connected crfs with gaussian edge
  potentials.
\newblock In {\em {NIPS}}.

\bibitem[\protect\citeauthoryear{Kwak, Hong, and Han}{2017}]{kwak2017weakly}
Kwak, S.; Hong, S.; and Han, B.
\newblock 2017.
\newblock Weakly supervised semantic segmentation using superpixel pooling
  network.
\newblock In {\em AAAI}.

\bibitem[\protect\citeauthoryear{Lin \bgroup et al\mbox.\egroup
  }{2014}]{lin2014microsoft}
Lin, T.-Y.; Maire, M.; Belongie, S.; Hays, J.; Perona, P.; Ramanan, D.;
  Doll\"ar, P.; and Zitnick, C.~L.
\newblock 2014.
\newblock Microsoft coco: Common objects in context.
\newblock In {\em {ECCV}}.

\bibitem[\protect\citeauthoryear{Long, Shelhamer, and
  Darrell}{2015}]{long2015fully}
Long, J.; Shelhamer, E.; and Darrell, T.
\newblock 2015.
\newblock Fully convolutional networks for semantic segmentation.
\newblock In {\em {CVPR}}.

\bibitem[\protect\citeauthoryear{Luc \bgroup et al\mbox.\egroup
  }{2016}]{luc2016semantic}
Luc, P.; Couprie, C.; Chintala, S.; and Verbeek, J.
\newblock 2016.
\newblock Semantic segmentation using adversarial networks.
\newblock {\em arXiv preprint arXiv:1611.08408}.

\bibitem[\protect\citeauthoryear{Noh, Hong, and Han}{2015}]{noh2015learning}
Noh, H.; Hong, S.; and Han, B.
\newblock 2015.
\newblock Learning deconvolution network for semantic segmentation.
\newblock In {\em {ICCV}}.

\bibitem[\protect\citeauthoryear{Pan \bgroup et al\mbox.\egroup
  }{2017}]{pan2017fully}
Pan, T.; Wang, B.; Ding, G.; and Yong, J.-H.
\newblock 2017.
\newblock Fully convolutional neural networks with full-scale-features for
  semantic segmentation.
\newblock In {\em AAAI}.

\bibitem[\protect\citeauthoryear{Papandreou \bgroup et al\mbox.\egroup
  }{2015}]{papandreou2015weakly}
Papandreou, G.; Chen, L.-C.; Murphy, K.~P.; and Yuille, A.~L.
\newblock 2015.
\newblock Weakly-and semi-supervised learning of a deep convolutional network
  for semantic image segmentation.
\newblock In {\em {ICCV}}.

\bibitem[\protect\citeauthoryear{Roy and Todorovic}{2017}]{roy2017combining}
Roy, A., and Todorovic, S.
\newblock 2017.
\newblock Combining bottom-up, top-down, and smoothness cues for weakly
  supervised image segmentation.
\newblock {CVPR}.

\bibitem[\protect\citeauthoryear{Russakovsky \bgroup et al\mbox.\egroup
  }{2015}]{russakovsky2015imagenet}
Russakovsky, O.; Deng, J.; Su, H.; Krause, J.; Satheesh, S.; Ma, S.; Huang, Z.;
  Karpathy, A.; Khosla, A.; Bernstein, M.; et~al.
\newblock 2015.
\newblock Imagenet large scale visual recognition challenge.
\newblock {\em {IJCV}} 115(3):211--252.

\bibitem[\protect\citeauthoryear{Saleh \bgroup et al\mbox.\egroup
  }{2016}]{saleh2016built}
Saleh, F.; Akbarian, M. S.~A.; Salzmann, M.; Petersson, L.; Gould, S.; and
  Alvarez, J.~M.
\newblock 2016.
\newblock Built-in foreground/background prior for weakly-supervised semantic
  segmentation.
\newblock In {\em {ECCV}}.

\bibitem[\protect\citeauthoryear{Shimoda and Yanai}{2016}]{shimoda2016distinct}
Shimoda, W., and Yanai, K.
\newblock 2016.
\newblock Distinct class-specific saliency maps for weakly supervised semantic
  segmentation.
\newblock In {\em {ECCV}}.

\bibitem[\protect\citeauthoryear{Simonyan and
  Zisserman}{2014}]{simonyan2014very}
Simonyan, K., and Zisserman, A.
\newblock 2014.
\newblock Very deep convolutional networks for large-scale image recognition.
\newblock In {\em {ICLR}}.

\bibitem[\protect\citeauthoryear{Souly, Spampinato, and
  Shah}{2017}]{souly2017semi}
Souly, N.; Spampinato, C.; and Shah, M.
\newblock 2017.
\newblock Semi and weakly supervised semantic segmentation using generative
  adversarial network.
\newblock {\em arXiv preprint arXiv:1703.09695}.

\bibitem[\protect\citeauthoryear{Wei \bgroup et al\mbox.\egroup
  }{2016}]{wei2016stc}
Wei, Y.; Liang, X.; Chen, Y.; Shen, X.; Cheng, M.-M.; Feng, J.; Zhao, Y.; and
  Yan, S.
\newblock 2016.
\newblock Stc: A simple to complex framework for weakly-supervised semantic
  segmentation.
\newblock {\em {IEEE TPAMI}}.

\bibitem[\protect\citeauthoryear{Wei \bgroup et al\mbox.\egroup
  }{2017}]{wei2017object}
Wei, Y.; Feng, J.; Liang, X.; Cheng, M.-M.; Zhao, Y.; and Yan, S.
\newblock 2017.
\newblock Object region mining with adversarial erasing: A simple
  classification to semantic segmentation approach.
\newblock In {\em {CVPR}}.

\bibitem[\protect\citeauthoryear{Zhou \bgroup et al\mbox.\egroup
  }{2016}]{zhou2016learning}
Zhou, B.; Khosla, A.; Lapedriza, A.; Oliva, A.; and Torralba, A.
\newblock 2016.
\newblock Learning deep features for discriminative localization.
\newblock In {\em {CVPR}}.

\end{thebibliography}
\bibliographystyle{aaai}
\end{document}